\documentclass[twoside]{article}

\usepackage[accepted]{aistats2024}

\usepackage{cite}

\usepackage{amsmath,amssymb,amsfonts}
\usepackage{algorithmic}
\usepackage{graphicx}
\usepackage{textcomp}
\usepackage{xcolor}
\usepackage{dsfont}
\usepackage{hyperref}
\usepackage{tikz}
\usepackage{pgfplots}
\usepackage{subcaption}
\usepackage{booktabs}
\usepackage{xspace}
\usepgfplotslibrary{colormaps}
\usepackage{adjustbox}

\usetikzlibrary{matrix}
\usepgfplotslibrary{groupplots}
\pgfplotsset{compat=newest}

\DeclareMathOperator{\cut}{\operatorname{Cut}}
\DeclareMathOperator{\ncut}{\operatorname{NCut}}

\DeclareMathOperator{\vol}{\operatorname{Vol}}
\DeclareMathOperator{\tr}{\operatorname{Tr}}
\newcommand{\specmix}{\textsc{SpecMix}\xspace}
\newcommand{\specmixonlycat}{\textsc{OnlyCat}\xspace}

\def\BibTeX{{\rm B\kern-.05em{\sc i\kern-.025em b}\kern-.08em
    T\kern-.1667em\lower.7ex\hbox{E}\kern-.125emX}}
\begin{document}

\twocolumn[

\aistatstitle{Spectral Clustering of Categorical and Mixed-type Data via Extra Graph Nodes}

\aistatsauthor{ Dylan Soemitro \And Jeova Farias Sales Rocha Neto}

\aistatsaddress{ \shortstack{Department of Computer Science \\ Columbia University} \And  \shortstack{Department of Computer Science \\ Bowdoin College}} ]




\begin{abstract}
Clustering data objects into homogeneous groups is one of the most important tasks in data mining. Spectral clustering is arguably one of the most important algorithms for clustering, as it is appealing for its theoretical soundness and is adaptable to many real-world data settings. For example, mixed data, where the data is composed of numerical and categorical features, is typically handled via numerical discretization, dummy coding, or similarity computation that takes into account both data types. This paper explores a more natural way to incorporate both numerical and categorical information into the spectral clustering algorithm, avoiding the need for data preprocessing or the use of sophisticated similarity functions. We propose adding extra nodes corresponding to the different categories the data may belong to and show that it leads to an interpretable clustering objective function. Furthermore, we demonstrate that this simple framework leads to a linear-time spectral clustering algorithm for categorical-only data. Finally, we compare the performance of our algorithms against other related methods and show that it provides a competitive alternative to them in terms of performance and runtime.
\end{abstract}


\section{Introduction}
Clustering is a central task in machine learning whose objective is to group objects according to a preestablished notion of group homogeneity, typically considering object similarity. Its application is widespread in many scientific fields, such as anthropology, biology, economics, marketing, and medicine \cite{clusteringBook}, which generally require its algorithms to be able to handle different types of data objects. Numerical data, for example, is composed of continuous features representing object information that lives in a spectrum, while categorical data handles features that are better represented as choices over a discrete set of options. In many practical scenarios, objects present both types of data in their feature set \cite{ahmad2019survey}, which motivates the research on methods that generalize numerical-only or categorical-only algorithms to this mixed-data setting.

The most intuitive generalization approach consists in preprocessing all features such that they all become fully categorical via discretization or numerical via dummy coding. Although these solutions are common in practice, it can be shown that they introduce distortion into the data and may lead to unsatisfactory clustering results \cite{foss2018kamila}. In view of this limitation, modern mixed-data methods attempt to directly involve both types of features in their algorithmic design and can roughly fall into one of two main categories \cite{preud2021head}. The first one establishes similarity metrics that involve both numerical and categorical information and uses them as part of an out-of-the-shelf similarity-based clustering \cite{huang1998extensions}. The second category, on the other hand, attempts to design holistic probabilistic models that account for both types and proceeds with Expectation Maximization-related methods to fit these models to the data, finding the underlying clusters along the way \cite{foss2016semiparametric, mcparland2016model}. 

Despite these methodological advances, only a few attempts have been made to extend spectral clustering (SC) to the mixed data setting. Due to the seminal work of Shi and Malik \cite{shi2000normalized}, when they proposed the Normalized Cuts criterion, this approach to clustering attracted popularity from theorists and practitioners alike. With SC, we have a simple algorithm with strong theoretical background \cite{meilua2001random, boedihardjo2021performance} that is adaptable to different settings \cite{chew2015semi, kleindessner2019guarantees} and is the method of choice in practice for graph data. However, for mixed-data clustering, the literature only provides us with three spectral solutions \cite{luo2006clustering, david2012spectralcat, mbuga2021spectral}, where the authors either apply SC to similarity matrices that take into account both numerical and categorical features, or initially discretize the numerical information so that the clustering is performed on purely categorical data.

In this paper, we demonstrate that one can incorporate the information carried by both numerical and categorical features into SC by \textit{structurally} changing the graph to be clustered. Here, we add sets of extra nodes to a base graph that is constructed using solely the available numerical data. Each node in the base is then connected to an extra node that corresponds to its categorical variable. We show that the discrete optimization that is approached by Normalized Cuts in this graph has a clear clustering interpretation that elegantly embodies the categorical information, naturally avoiding the necessity of discretization. To the best of our knowledge, this is the first instance of such a mathematically sound spectral approach in the mixed-type data clustering literature. Furthermore, we show that the proposed spectral framework is also ideal when dealing with purely categorical data \cite{qian2015space}, as we can then make efficient use of this graph structure to develop a linear clustering algorithm on the number of nodes. Finally, our experiments show that our algorithms outperform popular solutions to both mixed and categorical-only data in terms of both clustering quality and runtime.


\section{Related Work}

\subsection{Mixed-type Data Clustering}
In recent years, there has been an increasing interest in mixed data clustering, which has led to the development of numerous methods to tackle the problem. In a recent work, Preud’homme et al. \cite{preud2021head} identified two main categories for these algorithms: Model-based and Distance‐based (or partitional)\footnote{The reader is referred to \cite{ahmad2019survey} for a review of other approaches classified beyond these two main categories.}.

In model-based methods, each datapoint is assumed to match a predefined parametric statistical model, from which the clustering partition can be inferred. Here, analogously to distance-based methods (see below), the mixed-data setting is achieved by compromising two traditional mixture distributions for numerical and categorical data. In $K$-means for Mixed
Large Data (KAMILA) \cite{foss2018kamila}, for example, the authors suggest modeling the numerical and categorical features as arising from a spherical Gaussian Mixture Model (GMM) and form a Mixture of Multinomial (MM) distributions. The log-likelihood of the sum of these two components is used in a $K$-means algorithm setting to compute the final clusters. Clustering by Mixture Modeling (ClustMD) \cite{mcparland2016model} assumes a GMM and an MM model for the numerical and categorical features, respectively, and runs the standard Expectation Maximization (EM) algorithm to learn the parameters of the full model. Latent Class Analysis (LCA) \cite{weller2020latent}, on the other hand, only models categorical information, implying that continuous variables are discretized beforehand. Then, it assumes an MM model on the resulting dataset and proceeds with maximum likelihood via EM for parameter learning and clustering.

In distance-based methods, the main goal is to extend classical partitional clustering algorithms, such as the $K$-means algorithm \cite{lloyd1982least} to the mixed-data setting. $K$-prototypes algorithm \cite{huang1998extensions}, for example, alternate between (1) computing cluster centers represented by mean values for numeric features and mode values for categorical features, and (2) cluster assignments based on the minimal sum of Euclidean and Hamming distances from each datapoint to the identified cluster centers. Ahmad and Dey \cite{ahmad2007k}, on the other hand, improved $K$-prototypes' performance by replacing the Hamming distances measure for categorical datapoint dissimilarity with a cooccurrence-based metric. Other methods employ the strategy of first converting the mixed data to numeric data and then applying $K$-means on the new numeric data \cite{wang2015coupled, wei2015clustering}. In particular, Factor Analysis of Mixed Data (FAMD)\cite{pages2004analyse}, which generalizes both Principal Component Analysis (PCA) and  Multiple Correspondence Analysis (MCA) to the mixed-data setting, was also employed as a preprocessing dimensional reduction algorithm \cite{van2019distance} to be followed by $K$-means for clustering.

Within the class of partitional methods, one can also identify a small subset of algorithms that approach mixed-data Spectral Clustering (SC). To the best of our knowledge, there are three methods in this category and they approach the clustering problem by using SC on a similarity matrix that takes into account both numerical and categorical features. In \cite{mbuga2021spectral}, the Euclidean and Hamming distances are used to achieve the final pairwise similarity. In \cite{luo2006clustering}, the authors proposed a $K$-means-based discretization of the numerical features and then proceeded with the Hamming distance on the final dataset for the similarity computation. With SpectralCAT \cite{david2012spectralcat}, an optimal preprocessing discretization of numerical variables is performed instead of $K$-means, and then SC is applied to the resulting purely categorical data.

\subsection{Categorical Data Clustering}

$K$-modes algorithm \cite{huang1998extensions} is known to be one of the first algorithms to adapt the $K$-means methodology to purely categorical data. The algorithm alternates between computing cluster modes and Hamming distances to such modes. Due to its simplicity and practical success, many extended versions of $K$-modes were proposed  \cite{kim2005k, ng2007impact, bai2012impact, dinh2019estimating}, achieving various degrees of improvement over the original method. Their effort consisted mainly of employing a novel categorical similarity measure\footnote{The interested reader is referred to \cite{alamuri2014survey} for a review on similarity measures for categorical data.} \cite{ng2007impact, dinh2019estimating}, defining new cluster representatives \cite{san2004alternative, kim2005k, bai2012impact}, or optimally converting the original discrete categories into continuous entities to apply the standard $K$-means method \cite{qian2015space}. 

In addition to the $K$-modes algorithm and its variants, model-based solutions, such as LCA \cite{weller2020latent} (discussed above), hierarchical/agglomerative  \cite{ghattas2017clustering, he2002squeezer} methods have also been proposed. To the best of our knowledge, however, spectral clustering algorithms have not yet been designed to handle purely categorical datasets.

\section{Preliminaries}
\subsection{Notation}
For $n \in \mathbb{N}$, we have $[n] = \{1, \ldots, n\}$. $I_n$ is an identity matrix of size $n\times n$ and $0_{n \times m}$ is a matrix of zeros of size $n\times m$. For $B \in \mathbb{R}^{n \times n}$, $\tr(A) = \sum_{i \in [n]} B_{i, i}$ is its trace and $\lVert B \rVert_F = \sqrt{\tr(BB^{\top})}$ is its Frobenius norm with $(\cdot)^{\top}$ representing matrix transposition. The one-hot encoding $h \in \{0, 1\}^Q$ of a data object $p$ according to a categorical variable of $Q$ possible categories is a vector with entries $h(c) = \mathds{1}\{p \text{ belongs to the $c$-th category}\}, \forall c \in [Q]$, where $\mathds{1}\{\cdot\}$ is the indicator function. We use $|\cdot|$ to denote the size of a set and $[\cdot]$ for the concatenation of matrices (both row-wise and column-wise).

\subsection{Spectral Clustering}\label{sec:spec_clust}
Let $G = (V, E)$ be an undirected graph of $|V| = n$ nodes and edge set $E$. Let $W \in \mathbb{R}^{n \times n}$ be the matrix that encodes the positive similarities between the nodes. 
In Normalized Cuts ($\ncut$)\footnote{We only cover $\ncut$ for simplicity, although our methodology can be adapted to include Ratio Cuts \cite{von2007tutorial}.}, our goal is to find a partition $\{A_1, \ldots, A_K\}$ of $V$ in $K$ clusters, such as to minimize the following objective:
\begin{equation}\label{eq:ncut}
    \ncut(A_1, \ldots, A_K|G) = \sum_{k \in [K]} \frac{\cut(A_k, V \setminus A_k|G)}{\vol(A_k|G)},
\end{equation}
where we defined a cut as $\cut(A, B|G) = \sum_{i \in A, j \in B} W(i, j)$ and the volume of  a cluster is $\vol(A|G) = \sum_{i \in A, j \in V} W(i, j)$. 

Unfortunately, directly minimizing Eq. \ref{eq:ncut} is NP-hard \cite{von2007tutorial}, so we have to rely on approximations. Let $D \in \mathbb{R}^{n \times n}$ be the diagonal degree matrix of $G$, i.e., $D(i, i) = \sum_{j \in [n]} W(i, j), \forall i \in [n]$, and $L = D - W$  denote the graph Laplacian matrix of $G$. Let $X \in \mathbb{R}^{n \times K}$ be an \textit{assignment matrix} with entries:
\begin{equation}\label{eq:unrelax_X_def}
    Z(i, k) = 
    \begin{cases}
        \dfrac{1}{\sqrt{\vol(A_k)}}, & \text{if } i \in A_k \\
        0, & \text{otherwise}
    \end{cases}
\end{equation}
Minimizing $\ncut$ in Eq. \ref{eq:ncut} can be written as:
\begin{equation}\label{eq:ncut_unrelaxed}
\begin{aligned}
\min_{Z \in \mathbb{R}^{n \times K}} \quad & \mathcal{E}(Z|G) , 
\\ \textrm{s.t.} \quad & Z \text{ as defined in Eq. \ref{eq:unrelax_X_def}},   \\
\end{aligned}
\end{equation}
where $\mathcal{E}(Z|G) = \tr(Z^{\top}LZ)$ is what we define as the \textit{assignment energy} we wish to minimize. One can now  relax the above problem by replacing the requirement that $X$ should be of the form \ref{eq:unrelax_X_def}, with a weaker one that imposes that ${Z^{\top}DZ} = I_k$:
\begin{equation}\label{eq:ncut_relaxed}
\begin{aligned}
\min_{Z \in \mathbb{R}^{n \times K}} \quad & \mathcal{E}(Z|G) , \\
\textrm{s.t.} \quad & {Z^{\top}DZ} = I_k.   \\
\end{aligned}
\end{equation}
This new problem can be solved by (1) computing the matrix $V \in \mathbb{R}^{n \times K}$ whose columns are the eigenvectors $v_i, i\in [K]$ corresponding to the $K$ smallest eigenvalues  of the generalized eigenproblem  $Lv = \mu Dv$, and (2) applying $K$-means clustering \cite{lloyd1982least} to the rows of $V$.

\section{Proposed Methodology}

\subsection{Graph Construction and Algorithm}
Let $P = \{p^{(1)}, \ldots, p^{(n)}\}$ be a dataset of $n$ mixed data objects of $M$ features each, from which $R$ features are numerical and $Q$ are categorical.  Let $r^{(i)}_{\ell}$ and $q^{(i)}_{\ell}$ be the $\ell$-th numerical and categorical features of the $i$-th datapoint. 
Let $t_\ell$ be the number of categories from the $\ell$-th categorical feature and $t= \sum_{\ell \in [Q]} t_\ell$.

In our proposed graph, we initially create a base graph $G_{R} = (V_{R}, E_{R})$ with $|V_R| = n$  solely based on the numerical features of the points in $P$. From $G_R$ we create an augmented graph $G_{\text{all}}$ by adding $t$ extra nodes $\{F_{o, \ell}\}_{o \in [t_\ell], \ell \in [Q]}$, each corresponding to one category of the available categorical features. We then add $Q$ edges between each node $i$ in $V_R$ and the extra nodes that correspond to the categories listed in $q^{(i)}_{1}, \ldots, q^{(i)}_{Q}$. These extra edges will, respectively, have fixed weights $\lambda_{1}, \ldots, \lambda_{Q} > 0$. Figure \ref{fig:graph} depicts an example of $G_{R}$ with the added extra edges to form $G_{\text{all}}$. Our proposed method then proceeds with the usual spectral clustering algorithm on $G_{\text{all}}$, with $K$-means being applied to all nodes at the end. From that labeling, we only recover the labels for the nodes in $V_R$ for our final cluster assignments. We name this method \specmix.

\subsection{Cut Interpretation}
Let $h^{(i)}_{\ell}$ be the one-hot encoding vector of the $i$-th datapoint according to the $\ell$-th categorical variable and let $H_\ell \in \{0, 1\}^{n \times t_\ell}$ be a matrix with $h^{(i)}_{\ell}$ as its rows. The weight matrix $W_{\text{all}}\in \mathbb{R}^{(n + t)\times (n + t)}$ of $G_{\text{all}}$ is: 
\begin{equation}\label{eq:w_all}
    W_{\text{all}} = 
\begin{bmatrix}
W_R &  \lambda_1 H_1 &  \ldots  &  \lambda_Q H_Q \\
\lambda_1 H_1^\top & I_{t_1} &  \ldots  &  0_{n \times t_1} \\
\vdots & \vdots  &  \ddots  &  \vdots \\
\lambda_Q H_Q^\top & 0_{n \times t_1}  &  \ldots  &  I_{t_Q} \\
\end{bmatrix}.
\end{equation}
From $W_{\text{all}}$, one can compute the corresponding Laplacian matrix $L_{\text{all}} \in \mathbb{R}^{(n + t)\times (n + t)}$ as:
\begin{equation}
    L_{\text{all}} = 
\begin{bmatrix}
L_R + \lambda I_n &  -\lambda_1 H_1 &  \ldots  &  -\lambda_Q H_Q \\
-\lambda_1 H_1^\top & \lambda_1 D_{1}  &  \ldots  &  0_{n \times t_1} \\
\vdots & \vdots  &  \ddots  &  \vdots \\
-\lambda_Q H_Q^\top & 0_{n \times t_1}  &  \ldots  &  \lambda_{Q} D_{Q} \\
\end{bmatrix},
\end{equation}
where $\lambda = \sum_{\ell \in [Q]} \lambda_\ell$ and each $D_{\ell}$ is a diagonal matrix with $D_\ell(j, j) = \sum_{i \in [n]}H_\ell(i, j)$, i.e, the number of datapoints in $P$ that belong to category $j$ for the $\ell$-categorical variable. Now, assume we have a partition of the nodes in $G_{\text{all}}$. Let $X$ and $Y_1, \ldots, Y_Q$ be the assignment matrices for the nodes in $G_R$ and for the extra nodes, respectively, computed according to Eq. \ref{eq:unrelax_X_def} with respect to the current partition. Let $Z = [X\,\,Y_1\,\,Y_2\,\,\ldots\,\,Y_Q] \in \mathbb{R}^{n \times (n + t)}$. Then, the objective in Eq. \ref{eq:ncut_unrelaxed} becomes:
\begin{align}\label{eq:energy_z}
    \mathcal{E}(Z|G_\text{all}) =\nonumber & \tr(Z^{\top}L_\text{all}Z) 
    \\=\nonumber & \tr(X^\top (L_R +  \lambda I_n)X) 
    \\\nonumber&-  \sum_{\ell \in [Q]} \lambda_\ell \tr(2 (H_\ell Y_\ell)^\top X + Y_\ell^\top D_{\ell}Y_\ell)
    \\=  &\, \mathcal{E}(X|G_{R}) + \sum_{\ell \in [Q]} \lambda_\ell \lVert X - H_\ell Y_\ell\rVert^2_F,
\end{align}
where we used the fact that $D_{\ell} = H_\ell^\top H_\ell$. The first term in the above expression is the energy of the assignment $X$ when only considering the numerical information of the dataset.

To understand the impact of the remaining terms in Eq. \ref{eq:energy_z}, we first define $\delta_l(A, B)$ for the clusters $A$ and $B$  as the number of instances where a datapoint is clustered in $A$ and the node corresponding to its $\ell$-th categorical feature is assigned to $B$. Now, we notice that\footnote{When using Ratio Cut instead of Normalized Cut, this derivation assumes a cleaner form: $\lVert X - H_\ell Y_\ell\rVert^2_F = K - \sum_{k \in [K] } \delta_\ell (A_k, A_k)/|A_k|$.}:

\begin{figure}
    \centering
    \includegraphics[width=\linewidth]{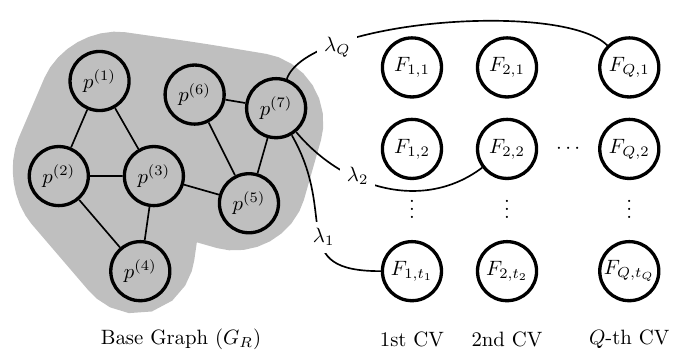}
    \caption{The proposed graph construction. Starting from a base graph computed solely using the available numerical variables, we add $t$ extra nodes corresponding to the categories of the available Categorical Variables (CV). Above, we depict how one of the nodes (the one corresponding to $p^{(7)}$) is connected to the extra nodes. In the above example, the point $p^{(7)}$  belongs to the categories represented by the nodes $F_{1, t_1}$, $F_{2, 2}$ and $F_{Q, 1}$.}
    \vspace{-10pt}
    \label{fig:graph}
\end{figure}

\begin{align}
    \lVert X - H_\ell Y_\ell\rVert^2_F &= \sum_{\substack{k, l \in [K] \\\nonumber k \neq l}}\frac{\delta_\ell (A_k, A_l)}{\vol(A_k|G_{\text{all}})} + \frac{\delta_\ell (A_k, A_l)}{\vol(A_l|G_{\text{all}})}
    \\\nonumber&= \sum_{k \in [K] } \frac{1}{\vol(A_k|G_{\text{all}})} \sum_{\substack{l \in [K] \\ k \neq l}}\delta_\ell (A_k, A_l)
    \\&= \sum_{k \in [K] } \frac{|A_k| - \delta_\ell (A_k, A_k)}{\vol(A_k|G_{\text{all}})}.
\end{align}
Therefore, finding $X$ and $Y_\ell$ that minimize $\lVert X - H_\ell Y_\ell\rVert^2_F$ effectively means to encourage partitions of the $n+t_\ell$ nodes in $G_{\text{all}}$ where each datapoint node is clustered together with the extra node related to its category for the $\ell$-th categorical variable. In other words, it penalizes clustering solutions where points in a given cluster belong to categories that are also those of points assigned to other clusters, instigating therefore each cluster to select a set of categories it should uniquely encompass.

In sum, minimizing the problem in Eq. \ref{eq:energy_z} has two effects: (1) it clusters our $n$ datapoints based on the similarities according to their numerical variables, while (2) encouraging that different categories across the points' categorical features are assigned to unique clusters. Therefore, \specmix approximates a cut solution that harmonically takes advantage of both numerical and categorical data without recurring to data type transformation or the formulation of a similarity function or statistical model that encompasses both. Besides that, this same framework can be used to design a fast and interpretable algorithm for purely categorical data.

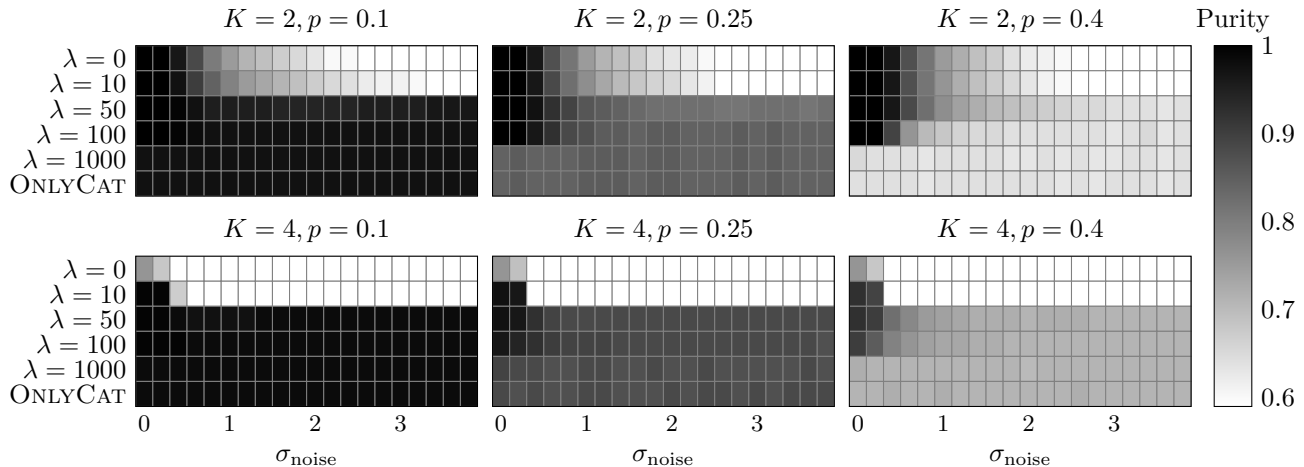
\begin{figure*}[!htb]
    \begin{tikzpicture}
        \begin{groupplot}[group style={group size= 3 by 2, vertical sep=0.8cm, horizontal sep=.2cm},height=2cm, width=0.265\textwidth,
        xticklabel style = {font=\footnotesize},
        scale only axis,
        ylabel near ticks,
        xlabel near ticks,
        xmin=-0.5, xmax=19.5,
        ymin=-0.5, ymax=5.5,
        title style={yshift=-1ex},
        point meta max=1,
        point meta min=.59,
        ytick style={draw=none},
        xtick style={draw=none},
        colormap={blackwhite}{color=(white) rgb255=(0,0,0)},
        ]
        \nextgroupplot[
            title={$K=2, p=0.1$},
            yticklabels={$\lambda = 0$, $\lambda = 10$, $\lambda = 50$, $\lambda = 100$, $\lambda = 1000$, \specmixonlycat}, 
            ytick=data,
            xticklabels={0, 0, 1, 2, 3, 4}, 
            xtick = {},
            xticklabels = {},
        ]
            \addplot[
            matrix plot,
            point meta=explicit, draw=gray
            ] table [meta=C] {data/vary_noise/scores_p1_k2_hm.dat};

        \nextgroupplot[
            title={$K=2, p=0.25$},
            ytick=data,
            xticklabels={0, 0, 1, 2, 3, 4}, 
            xtick = {},
            xticklabels = {},
            ytick = {},
            yticklabels = {},
        ]
            \addplot[
            matrix plot,
            point meta=explicit, draw=gray
            ] table [meta=C] {data/vary_noise/scores_p25_k2_hm.dat};

        \nextgroupplot[
            title={$K=2, p=0.4$ },
            ytick=data,
            xticklabels={0, 0, 1, 2, 3, 4}, 
            colorbar right,
            colorbar style={
                title=Purity},
            every colorbar/.append style={height=
            2*\pgfkeysvalueof{/pgfplots/parent axis height}+\pgfkeysvalueof{/pgfplots/group/vertical sep}},
            xtick = {},
            xticklabels = {},
            ytick = {},
            yticklabels = {},
        ]
            \addplot[
            matrix plot,
            point meta=explicit, draw=gray
            ] table [meta=C] {data/vary_noise/scores_p4_k2_hm.dat};

        \nextgroupplot[
            title={$K=4, p=0.1$},
            xlabel=$\sigma_{\text{noise}}$,
            yticklabels={$\lambda = 0$, $\lambda = 10$, $\lambda = 50$, $\lambda = 100$, $\lambda = 1000$, \specmixonlycat}, 
            ytick=data,
            xticklabels={0, 0, 1, 2, 3, 4}, 
        ]
            \addplot[
            matrix plot,
            point meta=explicit, draw=gray
            ] table [meta=C] {data/vary_noise/scores_p1_k4_hm.dat};

        \nextgroupplot[
            title={$K=4, p=0.25$},
            xlabel=$\sigma_{\text{noise}}$,
            ytick=data,
            xticklabels={0, 0, 1, 2, 3, 4}, 
            ytick = {},
            yticklabels = {},
        ]
            \addplot[
            matrix plot,
            point meta=explicit, draw=gray
            ] table [meta=C] {data/vary_noise/scores_p25_k4_hm.dat};

        \nextgroupplot[
            title={$K=4, p=0.4$},
            xlabel=$\sigma_{\text{noise}}$,
            ytick=data,
            xticklabels={0, 0, 1, 2, 3, 4}, 
            ytick = {},
            yticklabels = {},
        ]
            \addplot[
            matrix plot,
            point meta=explicit, draw=gray
            ] table [meta=C] {data/vary_noise/scores_p4_k4_hm.dat};

        \end{groupplot}
\end{tikzpicture}
    \vspace{-10pt}
    \caption{\specmix results on synthetic datasets when varying $\lambda$ . In all experiments, $n=1000$ and $Q=3$, and each line is the average of 50 synthetic experiments. All heatmaps share the same $y$-axis. The overall average runtimes are:  0.18s ($\lambda=0$), 0.24s ($\lambda=10$),  0.23s ($\lambda=50$), 0.23s ($\lambda=100$), 0.22s ($\lambda=1000$), 0.08s (\specmixonlycat).}\label{fig:vary_lambda}
\end{figure*}

\subsection{Case when dataset is categorical ($R = 0$)} \label{sec:onlycat}

Following the methodology described in Section \ref{sec:spec_clust}, our clustering algorithm relies on the solution of a generalized eigendecomposition on matrix $L_{\text{all}}$, which empirically requires $O(K(n+t)^2)$ operations to be performed if one is using the Lanczos method \cite{lanczos1950iteration}. Since $n \gg t$ in many practical settings practice \cite{uci2017}, our complexity is then similar to other similar mixed-type spectral clustering methods \cite{david2012spectralcat}.

Our method, however, attains a great computation advantage when our data is purely categorical, i.e., when $R = 0$ or $\sigma \to 0$. This is the case for many real datasets \cite{qian2015space, azen2021categorical, bai2022categorical} and for any methodology that converts numerical features into categorical data, such as SpectralCAT \cite{david2012spectralcat}. In this setting, $G_{\text{all}}$ effectively becomes a bipartite graph, which allows us to employ the Transfer Cut algorithm proposed in \cite{li2012segmentation}. Let $H = [\lambda_1 H_1\,\, \lambda_2 H_2\,\,\ldots\,\,\lambda_Q H_Q] \in \mathbb{R}^{n \times t}$.

Let $G_{Q} = (V_Q, E_Q)$ be a fully connected graph with $|V_{Q}| = t$ and its edge weights given by $W_{Q} = H^{\top} D_{H} H \in \mathbb{R}^{t \times t}$ where $D_{C}$ is diagonal with $D_{H}(i, i) = \sum_{j\in [t]} H(i, j) = Q$. Computing $W_{Q}$ takes $O(nt^2)$. Let $D_Q$ be the degree matrix of $G_Q$ and let $L_{Q}$ be the Laplacian matrix with respect to $G_{Q}$. 

Now, let $\{(\gamma_i, u_i)\}_{i \in [K]}$ with $0=\gamma_1 < \gamma_2 < \ldots < \gamma_k < 1$ be the eigensolution for $L_Q u = \gamma D_Q u$, and let $\{(\mu_i, v_i)\}_{i \in [K]}$ with $0=\mu_1 < \mu_2 < \ldots < \mu_k < 1$ be that for $L_{\text{all}} v = \gamma D_{\text{all}} v$. The authors in \cite{li2012segmentation} proved the following results:
\begin{equation*}
    \gamma_i (2-\gamma_i) = \mu_i, \quad f_i = \frac{1}{1-\gamma_i} D_H^{-1}H u_i, \quad v_i = \begin{bmatrix}
    f_i \\ u_i
    \end{bmatrix},
\end{equation*}
This means that the eigenvector decomposition for the whole node set, $\{(\mu_i, v_i)\}_{i \in [K]}$, can be efficiently inferred from the decomposition on $V_Q$, $\{(\gamma_i, u_i)\}_{i \in [K]}$. Considering that it takes $O(Kt^2)$ operations to compute $\{(\gamma_i, u_i)\}_{i \in [K]}$ (with Lanczos algorithm \cite{li2012segmentation}) and $O(Knt)$ to calculate $f_i = D_H^{-1}H u_i, i \in [K]$, we see that the total complexity of Transfer Cut in our setting is $O(nt^2) + O(Kt^2) + O(Knt) = O(nt(K+t) + Kt^2)$, which is linear in $n$. Once the eigenvectors $v_i$ are found, we only consider the first $n$ rows of the matrix $[v_1\,\, v_2\,\, \ldots\,\, v_k]$ in the subsequent $K$-means step. We refer to this algorithm as \specmixonlycat.

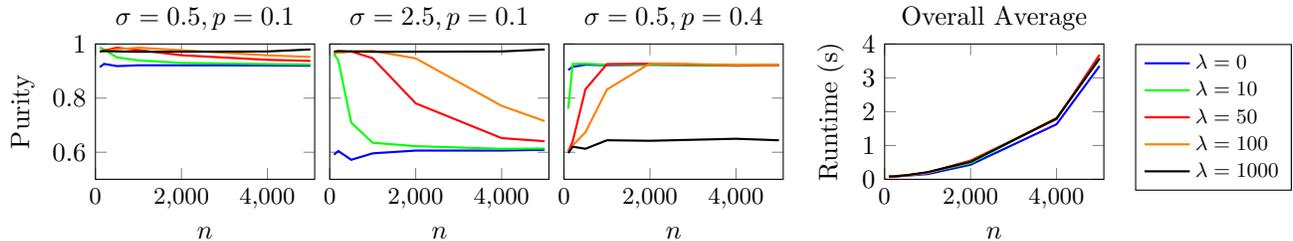
\begin{figure*}[!htb]
    \newcommand\len{3pt}
\newcommand\labelx{$\sigma$}
    \begin{tikzpicture}
        \begin{groupplot}[group style={group size= 3 by 2, vertical sep=.8cm, horizontal sep=.2cm},height=1.8cm, width=0.17\textwidth,
        xticklabel style = {font=\footnotesize},
        scale only axis,
        ylabel near ticks,
        xlabel near ticks,
        xmin=0,xmax=5100,
        ymin=0.5,ymax=1,
        title style={yshift=-1ex}
        ]
        \nextgroupplot[
            ylabel={Purity},
            xlabel={$n$},
            xtick = {},
            cycle list name=color list,
            title={$\sigma = 0.5, p=0.1$},
            ]
            \addplot[blue, thick] table[x=n, y=lambda_0_purity, col sep=comma] {data/vary_noise/var_n_sigma_05_p_01.csv};
            \addplot[green, thick] table[x=n, y=lambda_10_purity, col sep=comma] {data/vary_noise/var_n_sigma_05_p_01.csv}; 
            \addplot[red, thick] table[x=n, y=lambda_50_purity, col sep=comma] {data/vary_noise/var_n_sigma_05_p_01.csv}; 
            \addplot[orange, thick] table[x=n, y=lambda_100_purity, col sep=comma] {data/vary_noise/var_n_sigma_05_p_01.csv}; 
            \addplot[black, thick] table[x=n, y=lambda_1000_purity, col sep=comma] {data/vary_noise/var_n_sigma_05_p_01.csv}; 
            
        \nextgroupplot[
            xticklabel style = {font=\footnotesize},
            ytick = {},
            xlabel={$n$},
            yticklabels = {},
            xtick = {},
            title={$\sigma = 2.5, p=0.1$},
            cycle list name=color list,
            ]
            \addplot[blue, thick] table[x=n, y=lambda_0_purity, col sep=comma] {data/vary_noise/var_n_sigma_25_p_01.csv};
            \addplot[green, thick] table[x=n, y=lambda_10_purity, col sep=comma] {data/vary_noise/var_n_sigma_25_p_01.csv}; 
            \addplot[red, thick] table[x=n, y=lambda_50_purity, col sep=comma] {data/vary_noise/var_n_sigma_25_p_01.csv}; 
            \addplot[orange, thick] table[x=n, y=lambda_100_purity, col sep=comma] {data/vary_noise/var_n_sigma_25_p_01.csv}; 
            \addplot[black, thick] table[x=n, y=lambda_1000_purity, col sep=comma] {data/vary_noise/var_n_sigma_05_p_01.csv}; 

        \nextgroupplot[
            xticklabel style = {font=\footnotesize},
            xlabel={$n$},
            ytick = {},
            yticklabels = {},
            xtick = {},
            title={$\sigma = 0.5, p=0.4$},
            cycle list name=color list,
            ]
            \addplot[blue, thick] table[x=n, y=lambda_0_purity, col sep=comma] {data/vary_noise/var_n_sigma_05_p_04.csv};
            \addplot[green, thick] table[x=n, y=lambda_10_purity, col sep=comma] {data/vary_noise/var_n_sigma_05_p_04.csv}; 
            \addplot[red, thick] table[x=n, y=lambda_50_purity, col sep=comma] {data/vary_noise/var_n_sigma_05_p_04.csv}; 
            \addplot[orange, thick] table[x=n, y=lambda_100_purity, col sep=comma] {data/vary_noise/var_n_sigma_05_p_04.csv}; 
            \addplot[black, thick] table[x=n, y=lambda_1000_purity, col sep=comma] {data/vary_noise/var_n_sigma_05_p_04.csv};

    \end{groupplot}
\end{tikzpicture}%
\hspace{5pt}
\begin{tikzpicture}
        \begin{groupplot}[group style={group size= 1 by 1, vertical sep=.8cm, horizontal sep=.2cm},height=1.8cm, width=0.17\textwidth,
        xticklabel style = {font=\footnotesize},
        scale only axis,
        ylabel near ticks,
        xlabel near ticks,
        xmin=0,xmax=5100,
        ymin=0,ymax=4,
        title style={yshift=-1ex}
        ]
        \nextgroupplot[
            ylabel={Runtime (s)},
            xlabel={$n$},
            xtick = {},
            cycle list name=color list,
            title={Overall Average},
            legend style={nodes={scale=.87, transform shape}},
            legend style={at={(1.5,1)},anchor=north},
            legend cell align=left,
            legend style={font=\footnotesize}
            ]
            \addplot[blue, thick] table[x=n, y=lambda_0_time, col sep=comma] {data/vary_noise/var_n_sigma_05_p_01.csv};
            \addlegendentry{$\lambda=0$};
            
            \addplot[green, thick] table[x=n, y=lambda_10_time, col sep=comma] {data/vary_noise/var_n_sigma_05_p_01.csv};
            \addlegendentry{$\lambda=10$};

            \addplot[red, thick] table[x=n, y=lambda_50_time, col sep=comma] {data/vary_noise/var_n_sigma_05_p_01.csv}; 
            \addlegendentry{$\lambda=50$};

            \addplot[orange, thick] table[x=n, y=lambda_100_time, col sep=comma] {data/vary_noise/var_n_sigma_05_p_01.csv}; 
            \addlegendentry{$\lambda=100$};

            \addplot[black, thick] table[x=n, y=lambda_1000_time, col sep=comma] {data/vary_noise/var_n_sigma_05_p_01.csv}; 
            \addlegendentry{$\lambda=1000$};

    \end{groupplot}
\end{tikzpicture}
    \vspace{-17pt}
    \caption{\specmix results on synthetic data when varying the number of datapoints. In all experiments, $K=2$ and $Q=3$, and each line in the three left plots is the purity average of 50 synthetic experiments, while the one on the right is the average runtime of all experiments. All purity plots share the same $y$-axis.} \label{fig:vary_n}
\end{figure*}

\section{Numerical Experiments}
\subsection{Algorithmic and Experimental Setup}
In our experiments, for each pair of nodes $(i, j)$ in $G_R$ we computed the entries of the numerical weight matrix as $W_R(i, j) = \exp\left(-\sum_{\ell \in [R]}(r^{(i)}_{\ell} - r^{(j)}_{\ell})^2\right)$.
\noindent We decide on this fully connected graph formulation, instead of the more traditional $K$-NN graph \cite{von2007tutorial}, for simplicity. Also, for the sake of experimental simplicity, we set $\lambda_{1} = \ldots = \lambda_{Q} = \lambda$ in all of our experiments.

\begin{figure*}[!htb]
    \newcommand\len{3pt}
\newcommand\labelx{$\sigma$}
    \begin{tikzpicture}
        \begin{groupplot}[group style={group size= 3 by 2, vertical sep=.8cm, horizontal sep=.2cm},height=1.8cm, width=0.3\textwidth,
        xticklabel style = {font=\footnotesize},
        scale only axis,
        ylabel near ticks,
        xlabel near ticks,
        xmin=0,xmax=3.8,
        ymin=0.3,ymax=1,
        title style={yshift=-1ex}
        ]
        \nextgroupplot[
            ylabel={Purity},
            xtick = {},
            xticklabels = {},
            cycle list name=color list,
            title={$K=2, p=0.1$},
            ]
            \addplot[orange, dashed, thick] table[x=sigma, y=k-prototypes_purity, col sep=comma] {data/vary_noise/scores_p1_k2.csv}; 
            \addplot[green, dashed, thick] table[x=sigma, y=lca_purity, col sep=comma] {data/vary_noise/scores_p1_k2.csv}; 
            \addplot[blue, dashed, thick] table[x=sigma, y=spectralCAT_purity, col sep=comma] {data/vary_noise/scores_p1_k2.csv};
            \addplot[red, dashed, thick] table[x=sigma, y=famd_purity, col sep=comma] {data/vary_noise/scores_p1_k2.csv};
            \addplot[black, ultra thick] table[x=sigma, y=spectral_l50_purity, col sep=comma] {data/vary_noise/scores_p1_k2.csv};
            
        \nextgroupplot[
            xticklabel style = {font=\footnotesize},
            ytick = {},
            yticklabels = {},
            xtick = {},
            xticklabels = {},
            title={$K=2, p=0.25$},
            cycle list name=color list,
            ]
            \addplot[orange, dashed, thick] table[x=sigma, y=k-prototypes_purity, col sep=comma] {data/vary_noise/scores_p25_k2.csv};
            \addplot[green, dashed, thick] table[x=sigma, y=lca_purity, col sep=comma] {data/vary_noise/scores_p25_k2.csv}; 
            \addplot[blue, dashed, thick] table[x=sigma, y=spectralCAT_purity, col sep=comma] {data/vary_noise/scores_p25_k2.csv};
            \addplot[red, dashed, thick] table[x=sigma, y=famd_purity, col sep=comma] {data/vary_noise/scores_p25_k2.csv};
            \addplot[black, ultra thick] table[x=sigma, y=spectral_l50_purity, col sep=comma] {data/vary_noise/scores_p25_k2.csv};

        \nextgroupplot[
            xticklabel style = {font=\footnotesize},
            ytick = {},
            yticklabels = {},
            xtick = {},
            xticklabels = {},
            title={$K=2, p=0.4$ },
            cycle list name=color list,
            ]
            \addplot[orange, dashed, thick] table[x=sigma, y=k-prototypes_purity, col sep=comma] {data/vary_noise/scores_p4_k2.csv};
            \addplot[green, dashed, thick] table[x=sigma, y=lca_purity, col sep=comma] {data/vary_noise/scores_p4_k2.csv}; 
            \addplot[blue, dashed, thick] table[x=sigma, y=spectralCAT_purity, col sep=comma] {data/vary_noise/scores_p4_k2.csv};
            \addplot[red, dashed, thick] table[x=sigma, y=famd_purity, col sep=comma] {data/vary_noise/scores_p4_k2.csv};
            \addplot[black, ultra thick] table[x=sigma, y=spectral_l50_purity, col sep=comma] {data/vary_noise/scores_p4_k2.csv};

        \nextgroupplot[
            ylabel={Purity},
            xlabel = $\sigma$,
            xticklabel style = {font=\footnotesize},
            title={$K=4, p=0.1$},
            ]
            \addplot[orange, dashed, thick] table[x=sigma, y=k-prototypes_purity, col sep=comma] {data/vary_noise/scores_p1_k4.csv};
            \addplot[green, dashed, thick] table[x=sigma, y=lca_purity, col sep=comma] {data/vary_noise/scores_p1_k4.csv}; 
            \addplot[blue, dashed, thick] table[x=sigma, y=spectralCAT_purity, col sep=comma] {data/vary_noise/scores_p1_k4.csv};
            \addplot[red, dashed, thick] table[x=sigma, y=famd_purity, col sep=comma] {data/vary_noise/scores_p1_k4.csv};
            \addplot[black, ultra thick] table[x=sigma, y=spectral_l50_purity, col sep=comma] {data/vary_noise/scores_p1_k4.csv};

        \nextgroupplot[
            xlabel = $\sigma$,
            xticklabel style = {font=\footnotesize},
            ytick = {},
            yticklabels = {},
            title={$K=4, p=0.25$},
            legend columns=-1,
            legend style={at={(0.5,-.6)},anchor=north},
            legend style={cells={align=left}},
            legend style={nodes={scale=.87, transform shape}},
            ]
            \addplot[orange, dashed, thick] table[x=sigma, y=k-prototypes_purity, col sep=comma] {data/vary_noise/scores_p25_k4.csv};\addlegendentry{$K$-Prototypes};
            \addplot[green, dashed, thick] table[x=sigma, y=lca_purity, col sep=comma] {data/vary_noise/scores_p25_k4.csv}; \addlegendentry{LCA};
            \addplot[blue, dashed, thick] table[x=sigma, y=spectralCAT_purity, col sep=comma] {data/vary_noise/scores_p25_k4.csv};\addlegendentry{SpectralCAT};
            \addplot[red, dashed, thick] table[x=sigma, y=famd_purity, col sep=comma] {data/vary_noise/scores_p25_k4.csv};\addlegendentry{FAMD};
            \addplot[black, ultra thick] table[x=sigma, y=spectral_l50_purity, col sep=comma] {data/vary_noise/scores_p25_k4.csv};\addlegendentry{\specmix};

        \nextgroupplot[
            xlabel = $\sigma$,
            xticklabel style = {font=\footnotesize},
            ytick = {},
            yticklabels = {},
            title={$K=4, p=0.4$},
            ]
            \addplot[orange, dashed, thick] table[x=sigma, y=k-prototypes_purity, col sep=comma] {data/vary_noise/scores_p4_k4.csv};
            \addplot[green, dashed, thick] table[x=sigma, y=lca_purity, col sep=comma] {data/vary_noise/scores_p4_k4.csv}; 
            \addplot[blue, dashed, thick] table[x=sigma, y=spectralCAT_purity, col sep=comma] {data/vary_noise/scores_p4_k4.csv};
            \addplot[red, dashed, thick] table[x=sigma, y=famd_purity, col sep=comma] {data/vary_noise/scores_p4_k4.csv};
            \addplot[black, ultra thick] table[x=sigma, y=spectral_l50_purity, col sep=comma] {data/vary_noise/scores_p4_k4.csv};
            
            \coordinate (bot) at (rel axis cs:1, 0);

    \end{groupplot}
\end{tikzpicture}
\vspace{-15pt}
    \caption{Results on synthetic data for the tested methods. In all experiments, $n=1000$ and $Q=3$, and each line is the average of 50 synthetic experiments. All plots share the same $x$ and $y$-axis. The overall average runtimes are:  1.36s ($K$-prototypes), 0.16s (LCA), 1.06s (SpectralCAT), 0.11s (FAMD), 0.23s ($\specmix$).}
    \label{fig:synthetic_specmix}  
\end{figure*}
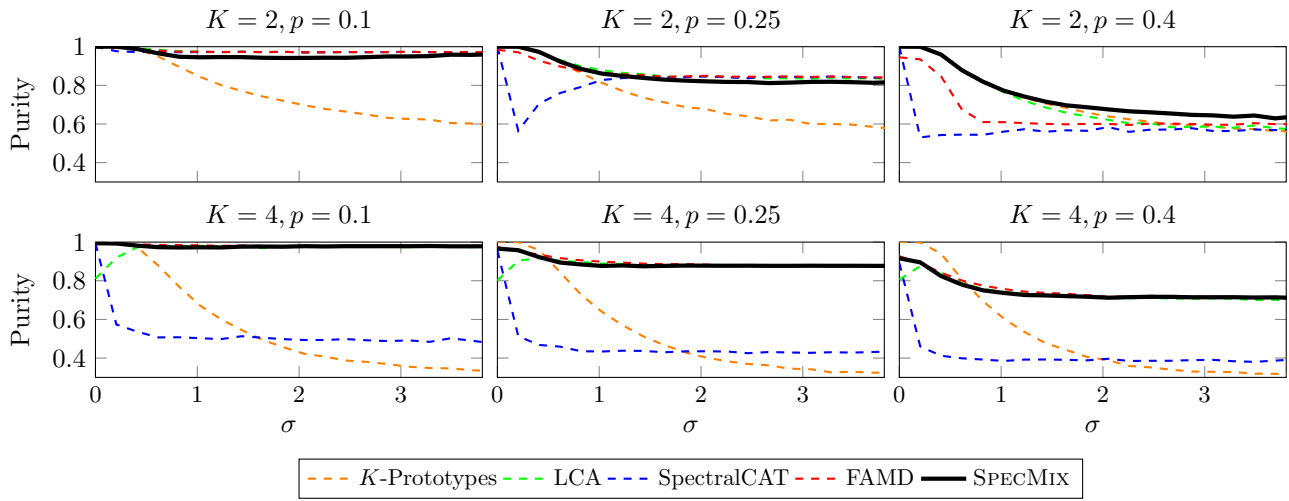
\begin{figure}[!h]
    \input{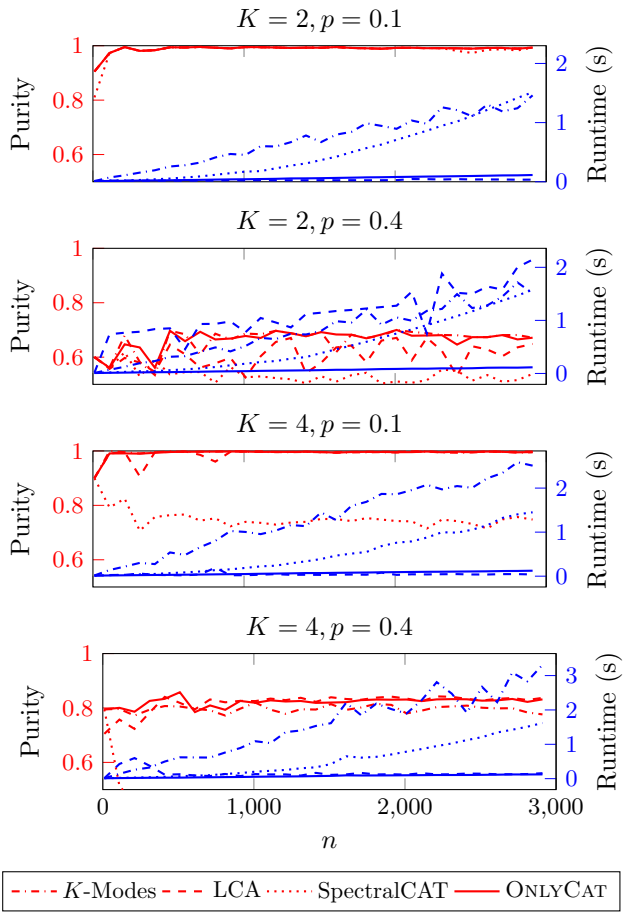}
    \vspace{-12pt}
    \caption{Results of synthetic categorical data for the tested methods. In all experiments, we set $Q = 5$. Each line is the average of 50 synthetic experiments. All plots share the same $x$-axis.}
    \label{fig:synthetic_onlycat}
\end{figure}

In our synthetic experiments, we generate a dataset with $K$ clusters and $n$ datapoints by sampling $n/K$ points in $\mathbb{R}^K$ from normal distributions with means given by the $K$ canonical basis vectors of that space ($[1, 0, 0]$, $[0, 1, 0]$ and $[0, 0, 1]$ if $K = 3$, for example) and the standard deviation $\sigma$, to be set in each experiment. To each of these (numerical) points, we add $Q$ categorical variables, with $K$ possible categories each. The categories for each point are chosen according to a value $p \in [0, 1]$ that quantifies how much each category is solely attached to a cluster. If $p = 0$, each category can only be found in one specific cluster. Otherwise, a category may be present in a cluster different from its attached cluster with probability $p$. For our experiments with real data, we use mixed type or categorical datasets available in the UCI Machine Learning Repository \cite{asuncion2007uci}. We consider ordinal variables in the data as also categorical and remove datapoints with missing values. We also keep track of the Imbalance Ratio of each dataset ($IR$), defined as the ratio between the sizes of the smallest and largest classes in the dataset \cite{mita2020libre}.

We compare \specmix with $K$-prototypes \cite{huang1998extensions}, LCA \cite{weller2020latent}, FAMD \cite{pages2004analyse}, KAMILA \cite{foss2018kamila} and SpectralCAT \cite{david2012spectralcat}, along with $K$-modes \cite{huang1998extensions} when using purely categorical datasets and applying the \specmixonlycat method. We selected these methods for their popularity in mixed-data clustering problems (the case of $K$-prototypes and KAMILA), their availability in a reliable implementation in Python or R, or their similarity to our proposed methods (SpectralCAT). We particularly make evident that using extra nodes in spectral clustering is more practically effective than the modeling proposed by SpectralCAT.
We follow the authors of SpectralCAT \cite{david2012spectralcat} and assess the performance of our algorithms according to the Purity Index (or simply Purity)\cite{tan2006introduction}. This measure considers the proportion of the majority ground-truth labels within each predicted cluster and averages that value across all clusters. 

We run our experiments on an AMD Ryzen 9 3900X 12-Core Processor, 3800 Mhz, 12 Cores with 32 GB RAM. Sample code with implementations of our methods will be available upon paper acceptance.

\subsection{Synthetic Data}
We start with the numerical experiments on synthetic datasets. In Figure \ref{fig:vary_lambda}, we evaluate our proposed methods in different scenarios for $\sigma$, $p$ and $K$, while $n = 1000$ is constant. For each data setting, we test \specmix under different values for $\lambda$, starting from $\lambda = 0$, which corresponds to the traditional spectral clustering algorithm applied only on the numerical data. We also added \specmixonlycat's performance for comparison. We notice that for large $\lambda$, \specmix effectively behaves as \specmixonlycat, which is expected since the numerical information amounts to little weight in this algorithmic regime. This demonstrates how $\lambda$ acts as smoothly transitioning \specmix from only considering the numerical information to only focus on their categorical parts.  This helps us to understand why the lower $\lambda$ is preferable when $\sigma$ (i.e., the corruption within the numerical data) is low, while large values of $\lambda$ reflect on better performances for lower $p$ (where the corruption within the categorical data is small). The main difference between $\specmix$ and $\specmixonlycat$ in the fully categorical scenario is speed, where $\specmixonlycat$ is preferable. From Figure \ref{fig:vary_lambda}, we can also recognize a ``sweet spot'' at around $\lambda=50$, where \specmix performs the best across all scenarios.

Figure \ref{fig:vary_n} depicts how the number of points in the dataset affects the performance of \specmix in terms of purity and runtime. We test our method in three different noise scenarios and five lambda values for $n \in \{200, 500, 1000, 2000, 4000, 5000\}$. Once again, $\lambda$ seems to smoothly transition the algorithm from purely numerical to purely categorical behavior. For low $p$ and high $\sigma$, our method tends to worsen its performance as $n$ increases, while the opposite happens in the high $p$ and low $\sigma$ scenario. This is potential evidence that numerical data still plays a more important role to the algorithm than categorical information in the $R = K = 2$ and $Q = 3$ case. Increasing $Q$ compared to $R$ should reverse this trend. The same figure also elucidates that the choice of $\lambda$ does not drastically affect the algorithm's runtime, which takes around 4s in the $n=5000$ setting.

Figure \ref{fig:synthetic_specmix} contrasts the performance of $\specmix$ against the other tested methods for datasets under various levels of difficulty according to the value of $\sigma$. In these experiments, we fix $\lambda=50$. We note that $\specmix$ is able to output cluster labelings that closely match or outperform the other methods in most scenarios, a phenomenon that is more explicit under higher $\sigma$. Overall, our proposed method is shown to be more reliable since it performs reasonably across all experiments, while all the other methods underperform in some data settings. Compared to SpectralCAT specifically, $\specmix$ is shown to better exploit the available categorical and numerical information when such is scarce, as seen when $\sigma$ and $p$ are large, greatly outperforming its spectral competitor. From Figure \ref{fig:synthetic_specmix}, we also note that $\specmix$, despite not being the quickest in runtime, is still competitive with the other methods. However, its computational efficiency could be further improved by sparsifying $G_{R}$ (by computing a $K$-nn graph, for example) since the remaining of $G_{\text{all}}$ is already sparse by construction.

        


In Figure \ref{fig:synthetic_onlycat}, we show the performance and timing of $K$-modes, LCA, and SpectralCAT against $\specmixonlycat$ for purely categorical synthetic datasets varying the size of the dataset. We excluded FAMD and KAMILA as they are not defined for this type of data. We fix $\lambda=1$ in these experiments, noting that any other value should produce the same results up to a numerical error. Here, our proposed method is shown to scale well with $n$, being able to attain competitive purity results in all dataset instances in under 0.3s at its slowest. In fact, except for LCA in low noise settings, $\specmixonlycat$ is the quickest and best-performing tested method. On close inspection, one also notices the linear growth in $\specmixonlycat$'s runtime, which is expected considering what is discussed in Section \ref{sec:onlycat}. These results not only demonstrate that $\specmixonlycat$ is a competitive methodology for clustering categorical data, but also suggest that using a graph formulation that embodies the categorical information via extra nodes is a performing and feasible solution for algorithmic scaling issues.

\begin{table*}[]

    \caption{Results on Real Datasets in terms of purity on real mixed-type (top rows) and categorical (bottom rows) datasets. Number of clusters and imbalance ratios are shown in parentheses. The best performance per dataset is shown in bold format, while the second best is underlined.}
    \label{tab:real}
    \begin{adjustbox}{max width=\textwidth}
    \setlength{\tabcolsep}{5pt}
        \begin{tabular}{ccccccc}\toprule
                 & \specmix  & $K$-prototypes & LCA & SpectralCAT & FAMD & KAMILA \\\cmidrule(lr){2-2}\cmidrule(lr){3-3}\cmidrule(lr){4-4}\cmidrule(lr){5-5}\cmidrule(lr){6-6}\cmidrule(lr){7-7}
        Student Dropout ($K = 2$, $IR = 0.35$)  & \underline{0.597}           & 0.546 & 0.511 & \textbf{0.643} & 0.553  & 0.529  \\
        Adult ($K = 2$, $IR = 0.33$)  & 0.751       & 0.751 & \textbf{0.772} & 0.751 & 0.751 & \underline{0.759} \\
        Hepatitis ($K = 2$, $IR = 0.19$)  & \underline{0.837}      & \underline{0.837}  & \textbf{0.850}  & \underline{0.837}  & \underline{0.837}  & \underline{0.837}  \\
        Post Operative ($K = 3$, $IR = 0.02$)  & \underline{0.724}       & \underline{0.724} & \textbf{0.735} & 0.712 & 0.712 & 0.713   \\
        Heart Disease ($K = 5$, $IR = 0.08$)  & \textbf{0.606}       & 0.538 & 0.582 & \underline{0.592} & 0.575 & 0.559\\
        Dermatology ($K = 6$, $IR = 0.18$)  & 0.779       & 0.349 & \underline{0.841} & \textbf{0.863} & 0.321 & 0.662 \\
        Zoo ($K = 7$, $IR = 0.09$)  & 0.772       & 0.821 & \textbf{0.891} & \underline{0.861} & 0.851 & 0.732 \\
        \midrule[1pt]
        \end{tabular}
    \end{adjustbox}
    
    \begin{adjustbox}{max width=\textwidth}
    \setlength{\tabcolsep}{16.5pt}
        \begin{tabular}{cccccc}
        
        & \specmixonlycat  & $K$-modes & LCA & SpectralCAT \\\cmidrule(lr){2-2}\cmidrule(lr){3-3}\cmidrule(lr){4-4}\cmidrule(lr){5-5}
        Lung Cancer ($K = 3$, $IR = 0.8$) & \underline{0.555}       & 0.518 & 0.518 & \textbf{0.629}   \\
        Car Evaluation ($K = 4$, $IR = 0.05$) & \underline{0.700}       & \textbf{0.708} & \underline{0.700} & \underline{0.700}   \\
        Soybean ($K = 19$, $IR = 0.25$) & \textbf{0.789}       & 0.609 & 0.722 & \underline{0.744}   \\
        Mushroom ($K = 23$, $IR = 0.61$) & \textbf{0.852}       & \textbf{0.852} & \textbf{0.852} &\underline{ 0.725}   \\
        \bottomrule
        \end{tabular}
    \end{adjustbox}
\end{table*}

\subsection{Real Data}
Table \ref{tab:real} depicts the clustering performance of each tested method on real datasets. We again refrain from applying FAMD and KAMILA on categorical datasets for the same reasons discussed in the previous section. We first standardize each numerical variable in each dataset before applying our proposed methods with parameters $\lambda=1$ for simplicity. Other values for $\lambda$ did not amount to greater overall performances.

Our methods have been shown to be competitive in some settings: low $K$/low $IR$ for \specmix and high $K$/high $IR$ for \specmixonlycat. In the case of mixed-data, we failed to outperform the other tested methods in some dataset instances, but always attained reasonable purity results. In particular, we are able to match or overperform KAMILA, the most popular mixed data clustering algorithm \cite{preud2021head}, in most cases. Despite being less principled than \specmix, SpectralCAT outperforms $\specmix$ in some instances, showing that the numerical discretization in their algorithm may be effective in practice.

Still, in Table \ref{tab:real} $\specmixonlycat$ is shown to outperform the tested methods in most of the tested datasets, especially those with higher $K$, again demonstrating its suitability for purely categorical data. Here, in particular, we note that these results were attained without parameter tuning. However, in practice, we noticed that other choices for $\lambda$ are more appropriate to specific datasets. Further improvements in both \specmix and \specmixonlycat could potentially be attained after a careful choice for the individual values of $\lambda_1, \ldots, \lambda_Q$, an endeavor that goes beyond the scope of this paper and is left to future work.

\section{Conclusion and Future Work}
In this work, we introduce \specmix, a spectral algorithm for the clustering of mixed data, and \specmixonlycat, for categorical data. Both methods rely on adding extra nodes, each corresponding to available data categories, to the traditional graph composed of datapoint-related nodes linked by their (numerical) similarities. Here, we show that this approach leads to an interpretable cut formulation that naturally incorporates the desired clustering behavior on both numerical and categorical data information. Effectively, the extra nodes act as a means of encouraging the clusters to encompass all points of individual categories, a desired outcome when considering categorical data. Algorithmically, we show that this framework is especially advantageous for purely categorical datasets when our proposed methodology can compute cluster assignments in linear time with respect to the number of datapoints. Our results show that our methods are competitive with many popular mixed and categorical data clustering algorithms.

For future work, we will explore how the embodiment of categorical data via extra nodes can affect the performance of other cut (such as Ratio-Cut) or diffusion-based clustering methods. We also hope to expand this concept to constrained clustering problems, as the extra nodes can incorporate data constraints that go beyond those investigated in this work. Moreover, we intend to develop strategies to automatically set our algorithm parameters ($\lambda_1, \ldots, \lambda_Q$) in real data and apply this methodology to settings where the numerical data might also have specific structural constraints, such as pixel data in image segmentation problems, where the base graph is structured as a grid.

\bibliographystyle{plain}
\bibliography{refs}

\begin{thebibliography}{10}

\bibitem{clusteringBook}
Charu~C. Aggarwal and Chandan~K. Reddy.
\newblock {\em Data Clustering: Algorithms and Applications}.
\newblock Chapman and Hall/CRC, 1st edition, 2013.

\bibitem{ahmad2007k}
Amir Ahmad and Lipika Dey.
\newblock A k-mean clustering algorithm for mixed numeric and categorical data.
\newblock {\em Data \& Knowledge Engineering}, 63(2):503--527, 2007.

\bibitem{ahmad2019survey}
Amir Ahmad and Shehroz~S Khan.
\newblock Survey of state-of-the-art mixed data clustering algorithms.
\newblock {\em Ieee Access}, 7:31883--31902, 2019.

\bibitem{alamuri2014survey}
Madhavi Alamuri, Bapi~Raju Surampudi, and Atul Negi.
\newblock A survey of distance/similarity measures for categorical data.
\newblock In {\em 2014 International joint conference on neural networks
  (IJCNN)}, pages 1907--1914. IEEE, 2014.

\bibitem{asuncion2007uci}
Arthur Asuncion and David Newman.
\newblock Uci machine learning repository, 2007.

\bibitem{azen2021categorical}
Razia Azen and Cindy~M Walker.
\newblock {\em Categorical data analysis for the behavioral and social
  sciences}.
\newblock Routledge, 2021.

\bibitem{bai2022categorical}
Liang Bai and Jiye Liang.
\newblock A categorical data clustering framework on graph representation.
\newblock {\em Pattern Recognition}, 128:108694, 2022.

\bibitem{bai2012impact}
Liang Bai, Jiye Liang, Chuangyin Dang, and Fuyuan Cao.
\newblock The impact of cluster representatives on the convergence of the
  k-modes type clustering.
\newblock {\em IEEE transactions on pattern analysis and machine intelligence},
  35(6):1509--1522, 2012.

\bibitem{boedihardjo2021performance}
March Boedihardjo, Shaofeng Deng, and Thomas Strohmer.
\newblock A performance guarantee for spectral clustering.
\newblock {\em SIAM Journal on Mathematics of Data Science}, 3(1):369--387,
  2021.

\bibitem{chew2015semi}
Selene~E Chew and Nathan~D Cahill.
\newblock Semi-supervised normalized cuts for image segmentation.
\newblock In {\em Proceedings of the IEEE International Conference on Computer
  Vision}, pages 1716--1723, 2015.

\bibitem{david2012spectralcat}
Gil David and Amir Averbuch.
\newblock Spectralcat: categorical spectral clustering of numerical and nominal
  data.
\newblock {\em Pattern Recognition}, 45(1):416--433, 2012.

\bibitem{dinh2019estimating}
Duy-Tai Dinh, Tsutomu Fujinami, and Van-Nam Huynh.
\newblock Estimating the optimal number of clusters in categorical data
  clustering by silhouette coefficient.
\newblock In {\em Knowledge and Systems Sciences: 20th International Symposium,
  KSS 2019, Da Nang, Vietnam, November 29--December 1, 2019, Proceedings 20},
  pages 1--17. Springer, 2019.

\bibitem{uci2017}
Dheeru Dua and Casey Graff.
\newblock {UCI} machine learning repository, 2017.

\bibitem{foss2016semiparametric}
Alex Foss, Marianthi Markatou, Bonnie Ray, and Aliza Heching.
\newblock A semiparametric method for clustering mixed data.
\newblock {\em Machine Learning}, 105:419--458, 2016.

\bibitem{foss2018kamila}
Alexander~H Foss and Marianthi Markatou.
\newblock kamila: clustering mixed-type data in r and hadoop.
\newblock {\em Journal of Statistical Software}, 83:1--44, 2018.

\bibitem{ghattas2017clustering}
Badih Ghattas, Pierre Michel, and Laurent Boyer.
\newblock Clustering nominal data using unsupervised binary decision trees:
  Comparisons with the state of the art methods.
\newblock {\em Pattern Recognition}, 67:177--185, 2017.

\bibitem{he2002squeezer}
Zengyou He, Xiaofei Xu, and Shengchun Deng.
\newblock Squeezer: an efficient algorithm for clustering categorical data.
\newblock {\em Journal of Computer Science and Technology}, 17(5):611--624,
  2002.

\bibitem{huang1998extensions}
Zhexue Huang.
\newblock Extensions to the k-means algorithm for clustering large data sets
  with categorical values.
\newblock {\em Data mining and knowledge discovery}, 2(3):283--304, 1998.

\bibitem{kim2005k}
Dae-Won Kim, KiYoung Lee, Doheon Lee, and Kwang~H Lee.
\newblock A k-populations algorithm for clustering categorical data.
\newblock {\em Pattern Recognition}, 38(7):1131--1134, 2005.

\bibitem{kleindessner2019guarantees}
Matth{\"a}us Kleindessner, Samira Samadi, Pranjal Awasthi, and Jamie
  Morgenstern.
\newblock Guarantees for spectral clustering with fairness constraints.
\newblock In {\em International Conference on Machine Learning}, pages
  3458--3467. PMLR, 2019.

\bibitem{lanczos1950iteration}
Cornelius Lanczos.
\newblock An iteration method for the solution of the eigenvalue problem of
  linear differential and integral operators.
\newblock 1950.

\bibitem{li2012segmentation}
Zhenguo Li, Xiao-Ming Wu, and Shih-Fu Chang.
\newblock Segmentation using superpixels: A bipartite graph partitioning
  approach.
\newblock In {\em 2012 IEEE conference on computer vision and pattern
  recognition}, pages 789--796. IEEE, 2012.

\bibitem{lloyd1982least}
Stuart Lloyd.
\newblock Least squares quantization in pcm.
\newblock {\em IEEE transactions on information theory}, 28(2):129--137, 1982.

\bibitem{luo2006clustering}
Huilan Luo, Fansheng Kong, and Yixiao Li.
\newblock Clustering mixed data based on evidence accumulation.
\newblock In {\em Advanced Data Mining and Applications: Second International
  Conference, ADMA 2006, Xi’an, China, August 14-16, 2006 Proceedings 2},
  pages 348--355. Springer, 2006.

\bibitem{mbuga2021spectral}
Felix Mbuga and Cristina Tortora.
\newblock Spectral clustering of mixed-type data.
\newblock {\em Stats}, 5(1):1--11, 2021.

\bibitem{mcparland2016model}
Damien McParland and Isobel~Claire Gormley.
\newblock Model based clustering for mixed data: clustmd.
\newblock {\em Advances in Data Analysis and Classification}, 10(2):155--169,
  2016.

\bibitem{meilua2001random}
Marina Meil{\u{a}} and Jianbo Shi.
\newblock A random walks view of spectral segmentation.
\newblock In {\em International Workshop on Artificial Intelligence and
  Statistics}, pages 203--208. PMLR, 2001.

\bibitem{mita2020libre}
Graziano Mita, Paolo Papotti, Maurizio Filippone, and Pietro Michiardi.
\newblock Libre: Learning interpretable boolean rule ensembles.
\newblock In {\em International conference on artificial intelligence and
  statistics}, pages 245--255. PMLR, 2020.

\bibitem{ng2007impact}
Michael~K Ng, Mark~Junjie Li, Joshua~Zhexue Huang, and Zengyou He.
\newblock On the impact of dissimilarity measure in k-modes clustering
  algorithm.
\newblock {\em IEEE transactions on pattern analysis and machine intelligence},
  29(3):503--507, 2007.

\bibitem{pages2004analyse}
Jerome Pages.
\newblock Analyse factorielle de donnees mixtes: principe et exemple
  d’application.
\newblock {\em Revue de statistique appliqu{\'e}e}, 52(4):93--111, 2004.

\bibitem{preud2021head}
Gregoire Preud’Homme, Kevin Duarte, Kevin Dalleau, Claire Lacomblez, Emmanuel
  Bresso, Malika Sma{\"\i}l-Tabbone, Miguel Couceiro, Marie-Dominique Devignes,
  Masatake Kobayashi, Olivier Huttin, et~al.
\newblock Head-to-head comparison of clustering methods for heterogeneous data:
  a simulation-driven benchmark.
\newblock {\em Scientific reports}, 11(1):1--14, 2021.

\bibitem{qian2015space}
Yuhua Qian, Feijiang Li, Jiye Liang, Bing Liu, and Chuangyin Dang.
\newblock Space structure and clustering of categorical data.
\newblock {\em IEEE transactions on neural networks and learning systems},
  27(10):2047--2059, 2015.

\bibitem{san2004alternative}
Ohn~Mar San, Van-Nam Huynh, and Yoshiteru Nakamori.
\newblock An alternative extension of the k-means algorithm for clustering
  categorical data.
\newblock {\em International journal of applied mathematics and computer
  science}, 14(2):241--247, 2004.

\bibitem{shi2000normalized}
Jianbo Shi and Jitendra Malik.
\newblock Normalized cuts and image segmentation.
\newblock {\em IEEE Transactions on pattern analysis and machine intelligence},
  22(8):888--905, 2000.

\bibitem{tan2006introduction}
Pang-Ning Tan, Michael Steinbach, and Vipin Kumar.
\newblock Introduction to data mining.
\newblock 2006.

\bibitem{van2019distance}
Michel Van~de Velden, Alfonso Iodice~D'Enza, and Angelos Markos.
\newblock Distance-based clustering of mixed data.
\newblock {\em Wiley Interdisciplinary Reviews: Computational Statistics},
  11(3):e1456, 2019.

\bibitem{von2007tutorial}
Ulrike Von~Luxburg.
\newblock A tutorial on spectral clustering.
\newblock {\em Statistics and computing}, 17:395--416, 2007.

\bibitem{wang2015coupled}
Can Wang, Chi-Hung Chi, Wei Zhou, and Raymond Wong.
\newblock Coupled interdependent attribute analysis on mixed data.
\newblock In {\em Proceedings of the AAAI Conference on Artificial
  Intelligence}, volume~29, 2015.

\bibitem{wei2015clustering}
Min Wei, Tommy~WS Chow, and Rosa~HM Chan.
\newblock Clustering heterogeneous data with k-means by mutual
  information-based unsupervised feature transformation.
\newblock {\em Entropy}, 17(3):1535--1548, 2015.

\bibitem{weller2020latent}
Bridget~E Weller, Natasha~K Bowen, and Sarah~J Faubert.
\newblock Latent class analysis: a guide to best practice.
\newblock {\em Journal of Black Psychology}, 46(4):287--311, 2020.

\end{thebibliography}

\newpage

\end{document}